# The Costly Dilemma: Generalization, Evaluation and Cost-Optimal Deployment of Large Language Models

Abi Aryan, Aakash Kumar Nain, Andrew McMahon, Lucas Augusto Meyer, Harpreet Singh Sahota

## Abstract

When deploying machine learning models in production for any product/application, there are three properties that are commonly desired. First, the models should be generalizable, in that we can extend it to further use cases as our knowledge of the domain area develops. Second they should be evaluable, so that there are clear metrics for performance and the calculation of those metrics in production settings are feasible. Finally, the deployment should be cost-optimal as far as possible. In this paper we propose that these three objectives (i.e. generalization, evaluation and cost-optimality) can often be relatively orthogonal and that for large language models, despite their performance over conventional NLP models, enterprises need to carefully assess all the three factors before making substantial investments in this technology. We propose a framework for generalization, evaluation and cost-modeling specifically tailored to large language models, offering insights into the intricacies of development, deployment and management for these large language models.

## 1    Introduction

According to two separate Gartner reports [1][2], 85% of AI and machine learning projects fail to deliver, with only 53% of projects finally making it from prototype to production. The four key reasons for this mentioned in a subsequent study by Gartner [3] were: 1) a lack of business-use case clarity, 2) inadequate skills within the team for end-to-end deployments, 3) neglecting organizational change, and 4) failure to experiment. These problems still remain in typical machine learning projects, and are becoming particularly acute as organizations attempt to adopt Large Language Models (LLMs).Further questions are important for those considering the adoption of LLMs such as the "build vs buy" hypothesis, should teams consume models hosted as third party services or build their own LLMs? How do the costs for LLM development, deployment and management scale once the business use-case has been established? And finally, given all the evaluation frameworks and leaderboards out there, how should engineering teams evaluate these models?

We assert that these challenges are particularly acute as organizations adopt and adapt LLMs for two main reasons associated with the ability to evaluate these models and deploy them cost effectively. First, compliance and other risks are high [4] due to lack of clear evaluation metrics, and secondly, there are hidden costs associated with model deployments for LLMs that must be accounted for. In this paper, we will explore these challenges and suggest some strategies for mitigating them.

The first part of the paper introduces the question of generalized vs. domain-specific LLMs,with a discussion of the advantages and disadvantages of reach through these risk and cost lenses. We then develop the analysis further in the second part as we cover the visible and hidden short and long-term costs of deploying these models in production. The third part focuses on how LLM Operations (LLMOps) techniques can help to tackle the aforementioned challenges.

Related research on the question of generalized vs. specialized technological solutions has been performed by E. Brynjolfsson et al. [5]. Rosa et. al [6] have performed cost-benefit analysis for multi-lingual language models, with a focus on one-time vs recurrent costs..

## 2 The GCE Trifecta

When developing new technology, questions of generalization, cost optimization, and evaluation, play a pivotal role in determining project success. Each component - generalization, cost, and evaluation - carries its own significance, yet they collectively contribute to achieving project objectives. We term these three elements the "GCE trifecta". The consideration of these components for the developments for LLMs is something that we believe has not been done in detail, we do this below.

Generalization stands as a fundamental pillar of LLM project success. It encompasses the ability of a large language model to deliver its intended outcomes across a broad range of contexts and situations. Generalization of the underlying model across different language based tasks enables scalability, adaptability, and the potential for replication, allowing projects to tackle complex challenges while remaining applicable to diverse scenarios.

Cost optimality serves as a critical factor in the ability of an organization to harness the potential of large language models. Accurate cost-benefit analyses of LLM enabled technological solutions will be needed to enable organizations to achieve their goals within budgetary constraints, maximizing value and gaining a return on investment.

Lastly, the ability to evaluate models acts as the cornerstone of machine learning project success. Evaluation provides a systematic and objective assessment of the model performance across likely production scenarios. By employing rigorous evaluation methodologies, product teams can ensure accountability, transparency, and the ability to do continuous experimentation and improvement throughout the lifecycle of their solution. Effective evaluation techniques enable stakeholders to gauge the impact of the LLM and make data-driven decisions for future endeavors. Consistent and robust evaluation methodologies and frameworks need to be used by organizations looking to adopt LLMs in order to enable these benefits and to mitigate the risks of failure in deployment scenarios.

Despite their inherent interdependencies, we propose that, for LLMs, the concepts generalization, cost optimization, and evaluation are relatively orthogonal in the context of project success. By recognizing the challenges within each of these areas and employing tailored strategies for specific use-cases, organizations can strike an appropriate balance, and create value through lasting project success. Through this research, we aim to shed light on the unique dynamics of the GCE trifecta for LLMs and provide insights that would be helpful across the organization and different stakeholders within the team.

## 2.1    Generalization

Broadly speaking, there are two different interest-groups amongst enterprises working on large language models. The first are the Foundation Model (sometimes also referred to as Base Model) providers. These providers aim to make it easy for anyone to access pre-trained large language models using cloud-based or self-hosted infrastructure. Depending on the provider, these models may be open-source or proprietary, based on the release strategy [7] of the provider. Amongst this category are companies like OpenAI, Cohere, Google, Microsoft, Anthropic, Nvidia, Mosaic, Hugging Face and others. While the out-of-the-box direct use of a foundational model may be the quickest way to deploy a LLM-based product or application. However, it may not add much substantial value depending on the use-case. Some organizations may instead benefit from domain-specific knowledge injection to improve task-specific performance of the models. While LLMs have seen adoption for several NLP applications across a wide-range of industries including coding assistants, copywriting, language prompted visual design, drug discovery, legal reviews and many others, an extensive review of all existing applications across industries is missing. That said, several economists have done reviews on the potential impact of GPTs on the labor market [12][13][14]. These reports suggest that LLMs are as of now more generally used for generative use-cases than discriminative use-cases. At the time of writing, we note some of the most popular LLM use-cases are knowledge retrieval [8], recommender systems [9], cross-lingual translation [10] and autonomous agents or AI Agents [11].

In many of these applications, the suggested mechanism for using LLMs is to consume them via a third party service. However, using these models out of the box exposes organizations to several new risks including those around regulatory and legal compliance of generative models trained on datasets with unknown provenance, security risks through new attack vectors like prompt injection and risks of drop in performance due to prompt drifts to name just a few. Thus, the trend of organizations fine-tuning their models on their industry-specific data, or even building them in-housem will likely accelerate. The ability to do this depends on factors such as the business use-case maturity and the capability of the technical teams within these entities.

The generalizability of these models is also of interest to developers and teams interested in integrating LLMs into their own, existing,products and services. These include Plugins, AI Agents as well as products that use LLMs for NLP applications, for tasks like summarization.

This can be challenging as at the time of writing there are no substantive studies that compare which provider would be better for which particular use-case, as the majority of benchmark cases focus on general applicability of the models or performance of only indirect relevance to many use cases. Another important factor that could guide the technical choices of different organizations in this space will be in terms of which models can be generally useful but still respect legislation(eg. the EU-AI Act [15]).

## Grading Foundation Model Providers' Compliance with the Draft EU AI Act

Source: Stanford Research on Foundation Models (CRFM), Institute for Human-Centered Artificial Intelligence (HAI)

| Draft AI Act Requirements | OpenAI GPT-4 | Cohere Command | Stable Diffusion v2 | Claude | PaLM 2 | BLOOM | LLaMA | Jurassic-2 | Luminous | GPT-NeoX | Totals |
|---|---|---|---|---|---|---|---|---|---|---|---|
| Data sources | | | | | | | | | | | 22 |
| Data governance | | | | | | | | | | | 19 |
| Copyrighted data | | | | | | | | | | | 7 |
| Compute | | | | | | | | | | | 17 |
| Energy | | | | | | | | | | | 16 |
| Capabilities & limitations | | | | | | | | | | | 27 |
| Risks & mitigations | | | | | | | | | | | 16 |
| Evaluations | | | | | | | | | | | 15 |
| Testing | | | | | | | | | | | 10 |
| Machine-generated content | | | | | | | | | | | 21 |
| Member states | | | | | | | | | | | 9 |
| Downstream documentation | | | | | | | | | | | 24 |
| Totals | 25 / 48 | 23 / 48 | 22 / 48 | 7 / 48 | 27 / 48 | 36 / 48 | 21 / 48 | 8 / 48 | 5 / 48 | 29 / 48 | |

Figure 1: *Stanford HAI evaluation of foundation model providers for their compliance with proposed EU law on AI.*

Such regulatory guardrails may potentially limit the availability of a certain model or provider by the region depending on the compliance-levels, see Figure 1). On a more technical level, deciding which provider and foundational model would be a better choice depends not only on the generalizability of the model, but also on several factors including number of parameters, size of context window, training type, inference speed, cost, fine-tunability as well as data security.

While there are subjective quality-measures for all models, the extensive model quality depends on your particular domain as well as the sophistication of application using a LLM. For example, if the goal is to do knowledge retrieval on unstructured data, GPT-4 may be an excellent choice however when doing knowledge retrieval on structured data, a model with a larger context window may be the most optimal choice. Although, one of the key limitations with building using proprietary models is there is a lack of best-practices and information on long-term-support (LTS), prompt-drift on updates and other important operational factors. This lack of clarity can lead to operational risks around reliability, explainability and predictability when moving into production.

Taken together, these points show that the selection of a model cannot be based solely on its ability to generalize across problem domains, and must be more nuanced in terms of applicability to organizational problems.

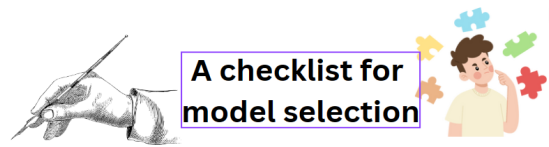

**A checklist for model selection**

- Is the Open-Source/Proprietary? (model parameters available or not?)
- What are the number of the parameters that the model does have?
- Is the training data available?
- What does the inference speed/latency of this model look like?
- What is the cost of inferencing this model?
- Is the model fine-tunable?
- What is the cost of fine-tuning this model?
- What is the size of the context-window for the model?

## 2.2    Cost Optimality

A very pertinent question that concerns every organization looking to deploy or utilize LLMs today is, how much will this cost?

First, consider building an in-house LLM and maintaining it in production. This is no easy feat and may require a significant investment in infrastructure, data collection, and hiring skilled personnel. As in other software deployments this will not be a one-time investment and will require ongoing operational cost. The full cost profile of LLMs is currently unknown and requires future work and data to determine. The uniqueness of the challenge for LLMs comes from the fact that the infrastructure and MLOps requirements for running such large and complex models are not common experience among the community yet. We expand on these challenges in the paragraphs below.

In contrast, vendor-based LLMs like GPT-4 may seem easier to estimate costs for, given they operate on a pay-as-you-go model. This can reduce upfront costs and allow for better operational cost modeling, making them an attractive choice.

**Table 1. Build vs Buy Hypothesis**

| Attribute | Build | Buy |
|---|---|---|
| Predictable Workload | Native LLMs require significant upfront capital investment for hardware, software, networking equipment, and facilities.<br><br>Ongoing costs include maintenance, staff, and energy consumption. | Vendor-Based LLMs typically operate on a pay-as-you-go model, reducing upfront costs and allowing for better cost control.<br><br>They provide almost limitless scalability, enabling organizations to easily expand or contract their |
| | | resources according to demand |
| Hardware Costs | Upfront and maintenance | Nominal |
| Security | More secure | Depends on the provider's infrastructure |
| Compliance | High Compliance | Check Chart (on Pg.2) |
| Latency | Lower Latency | Higher Latency |

This build-vs-buy dichotomy is not anything new in software development or in MLOps, but as we have mentioned it does have some new dimensions to consider for LLMs. Table 1 starts to flesh out some of these points.

Deploying an LLM is very different from deploying any other machine learning model because the cost in the case of LLMs is two-fold: infra-model related cost, and the hidden cost.

Although we can significantly reduce our costs with these vendor-based models, they come with their own sets of challenges. Organizations may be reticent when it comes to sharing sensitive data with any API-based LLM vendor. Organizations will also require that vendor based models are compliant with their own organizational data policies. This can be hard to ensure if data and implementation details are not shared freely.

Given this, there is the risk of lock-in as switching to a new model or vendor will incur new operational costs as these due diligence and compliance exercises are completed. In short, security, compliance, and latency become major concerns when choosing a vendor-based model. This can be more difficult to factor into a direct cost comparison, but must be considered,

Another dimension to consider when working with LLMs is the question of prompt engineering vs. fine-tuning. This choice depends on several factors, and has some associated consequences on the overall cost. An important consideration for this question is the length of the context window of the model and understanding how the overall cost is related to it. Although transformer models show exceptional scaling capabilities, one computational bottleneck that remains an open-challenge is the processing of long sequences. The complexity of the naive attention mechanism grows quadratically in terms of both the compute and the memory [17].

With the latest Anthropic model, we have a context window of 100K tokens that translates to roughly 75,000 words. Such a lengthy context window opens up opportunities for accomplishing tasks that were almost impossible to achieve in the past. For example, you can input an entire book into the model and dynamically query the content just from that provided context. This is a qualitative step up compared to the capabilities of some earlier LLMs.

With the larger context windows, you can retain more in-context information and the model can handle more complex and longer inputs. However, one of the challenges of large context windows is that the costs increase almost quadratically as the number of tokens are increased and can also affect the inference latency due to the slow-down of model computations. For example, Anthropic latest model response time on a 100K context window is roughly 22 seconds. Also, most use cases don't require such a large context window. It is also not so easy to write and modify lengthy prompts.

Smaller context windows allow for smaller input lengths thus requiring clear, concise, and clever prompts to obtain a desirable output. One of the advantages of short prompts is that they are easy to write and modify compared to the lengthier prompts. The overall latency is low, the chain of thoughts becomes easy and they also enable faster iteration. You can also leverage parallel context windows for many use cases to achieve acceptable performance on a task [18] without fine-tuning or using an expensive model with a bigger context window.

While it seems like there is an obvious upside to using smaller context windows and investing your time in prompt engineering, the iterative costs of this can accumulate quickly. To obtain a similar result from an LLM, you may be required to write multiple prompts and make multiple calls to the model. With multiple prompts and calls, this quickly adds to your overall inference cost of the model.

Another disadvantage of shorter prompts is that it makes it hard to decide when to go with fine-tuning instead of prompting. Most of the time prompting can take you far, but it may require you to run several trials before you decide to fine-tune, either with an explicit reward function or Reinforcement Learning from Human Feedback (RLHF).

Fine-tuning is substantially more expensive than prompting and not always the right approach depending on the complexity of the conditional. There may exist valid inferences that satisfy the conditional argument, however sampling them can be incredibly hard if we don't already know the factorization ahead of time.

Generally speaking, for domain-adaptation, prompt engineering works well for embedding-based search, fine-tuning may produce better results for categorization and filtering however there is no conclusive work that compares the generalization for both the options.

Another factor to keep in mind are the scaling behavior of these models.It has been proven empirically that language models get predictably better as the number of parameters, amount of compute used, and dataset size increases [16][19]. Query caching can significantly reduce costs for LLM inferencing, but the exact amount of cost reduction depends on various factors, such as the frequency of repeated queries, performance of the underlying language model, cache hit ratio, and the overall architecture of the LLM inferencing system. [20].

Once again, we hope that these points highlight some of the nuances of considering the cost profile of working with LLMs.

While the above discussion focused on the infrastructure and the modeling-related costs, LLMs also have associated hidden costs.

The first key cost is the potential cost of prompt drift. LLMs offer very little reproducibility even with the same prompts between different versions of the model. This often occurs as the model is updated to a faster, better, distilled LLM. Second, with traditional machine learning models, it is common to hire annotators to annotate datasets. Hiring annotators is cheap, and it takes a very short amount of time to train them for the defined annotation task. Validating the annotations done by the annotators is easy, and we can in many cases even automate the validation process to a large extent.

However, the same process becomes very complex in the case of LLMs due to the need for domain-specific knowledge to create good prompts. Either the relevant team writes and validates all the prompts every time, or you hire a prompt engineer. Hiring prompt engineers is expensive. On top of that, it creates an indirect dependency on the prompt engineer within the team if they choose to leave. Retraining new prompt engineers for your tasks is time-consuming, and expensive (see the discussion on API call cost accumulation above). Even if you hire a prompt engineer, there is currently no way to automate the validation of prompts.

Given that you can choose to call the model from the front-end or the back-end, and fine-tune vs. prompt engineer, this can result in operational costs that vary across a wide scale. LLMs are still relatively new in the machine learning world, which means that there are unknowns associated with using them in production. Some typical risks associated with machine learning models when used in production are listed below, along with a brief discussion of the important points regarding LLMs:

**- Compliance and regulatory risk**: This refers to the risk of breaking rules set by governing bodies of various flavors, be they government themselves, regulators or other institutions with powers to enforce compliance with set rules. In this scenario organizations can face potential large fines or other punitive measures. For example the upcoming EU AI Act, which is undergoing final review by European lawmakers at the time of writing, could mean fines of 10 million euros or 2% of global profits (whichever is higher) on organizations that breach these rules [21].

**- Reputational risk**: A system may not necessarily breach legal or regulatory guidelines but it may still behave outside the expected norms for interaction with a variety of stakeholders. Some examples could be a banking customer being faced with derogatory remarks, a hospital patient being blamed for their illness, or a customer being given suggestions that conform to racial or gender biases [22]. These scenarios can then lead to huge reputational damage for organizations and for the concept of AI and ML systems as a whole and lead to losses in income and future revenue.

**- Operational risk:** Many organizations now use data and machine learning to inform operational decisions. If an LLM generates inconsistent output or leads to an erroneous operational decision this could incur large costs as well. For example, if an LLM based chatbot was being used by a senior executive in an organization to help make an investment decision, hallucinated facts could lead to a large amount of that investment being wasted in a low growth area. Similarly incorrect information may lead to erroneous decisions around technology adoption, system design, logistical operations, administration execution that could lead to very costly outcomes.

**Table 2. Costs Associated with LLM Applications**

| Upfront Costs | Hidden Costs |
|---|---|
| Context Window | Model Drift, Prompt Drift |
| Prompting/Fine-Tuning | Hardware Costs |
| Data | Compliance |
| Infrastructure | People |
| Scalability | Reliability |

## 2.3    Evaluation

With the rise of Large Language Models (LLMs) the question of what best evaluation practice looks like must be revisited as some assumptions usually employed for ML evaluations no longer hold and need augmented. Large language models are often trained on massive amounts of data and require more than a few million parameters further limiting their reproducibility as well as interpretability.

This gets even more tricky as more and more companies are moving to closed-version of the models keeping the model parameters as well as information on RLHF and red-teaming the models through adversarial examples private thus making it close to impossible to fully evaluate these models.

In the past, transformer based language models were typically evaluated using perplexity, the BLEU score and Human Evaluations.[23] However, these metrics have been criticized for being too simplistic and not taking into account much of the nuance of human language. This is counteracted somewhat when using techniques based on human evaluation, however this can also be the most time-consuming and expensive approach, with particular challenges around scaling to large input and outputs, as is the case with LLMs.

There are several general benchmarks available for LLMs, namely OpenAI Evals, HELM, Evals-Harness, etc. however elo-based systems [25] are quickly gaining popularity over community-based leaderboards [24] v/s vendor-based evals (Nemo Guardrails, Aviary, etc). While the above-mentioned generalized benchmarks are helpful for some contexts, most organizations need domain-specific benchmarks that are specific to the company and their business use-case.

We break it down into five concerns that need to be addressed to develop a comprehensive evaluation framework.

| | HELM / lm-evaluation-harness | OpenAI/eval | Alpaca Evaluation | Vicuna Evaluation | Chatbot Arena |
|---|---|---|---|---|---|
| Question Source | Academic datasets | Mixed | Self-instruct evaluation set | GPT-4 generated | User prompts |
| Evaluator | Program | Program/Model | Human | GPT-4 | User |
| Metrics | Basic metrics | Basic metrics | Win rate | Win rate | Elo ratings |

Figure 2. LMSys.org - Comparison between different evaluation methods

### 2.3.1 What does "performance" mean for an LLM application?

Since LLMs do not have clear objective functions, thus it is hard to conventionally evaluate them using the conventional ML metrics. Thus, performance comes down to a combination of several factors-

1. Accuracy
2. Inference Speed
3. Latency

### 2.3.2 How do we create stable experimental setups for evaluating LLM applications?

While having a test set you test/benchmark against is incredibly important. However, the setup comes down to -

1. Data Sampling in Test vs Evals
2. Logging Prompts and Inferences
3. Checkpointing the models.

### 2.3.3 What benchmarks do we have or do we need to create to enable consistency of approach?

The generalized benchmarks depending on the use-case do allow for grounding. However, organizations still standard proxies like accuracy and other metrics against your own test dataset and/or public benchmarks.

### 2.3.4 For third party hosted models, what assurances can we give ourselves as downstream consumers through validation?

Delegating the model development and maintenance to vendors like OpenAI, Anthropic etc does allow one to have significant assurances when it comes to model staling, latency, scalability and easy deployment.

### 2.3.5 How do we evaluate and monitor the accuracy of our LLM-based solutions during development and post-deployment?

For applications, public benchmarks are not useful because it's not measured on the data distribution you care about (data your users give). So building elo-based benchmarks for your data can be an important step in the right direction.

## 3. Conclusion

The integration of Language Models (LLMs) into applications brings forth numerous benefits, but it also introduces the concept of technical debt. This debt can manifest as potential risks or hidden costs that may arise in the future. However, it is important to emphasize that LLMs remain highly valuable despite these considerations, and technical debt itself is not inherently negative. Just as individuals make informed decisions regarding financial debt and actively manage it, a similar approach must be adopted when dealing with technical debt in LLM-based solutions.

Choosing an appropriate level of technical debt becomes crucial in LLM integrations. This involves carefully evaluating the trade-offs between short-term gains and long-term consequences. LLMs offer immediate advantages such as enhanced natural language processing capabilities and improved user experiences. However, hasty implementation or overreliance on LLMs without addressing potential technical debt can lead to challenges down the line.

Managing technical debt in LLM-based solutions requires a proactive and strategic approach. Just as financial debt requires diligent monitoring and repayment plans, technical debt should be assessed, documented, and accounted for. Organizations must invest resources in identifying areas where technical debt may accumulate, such as code complexity, potential performance bottlenecks, or lack of maintainability. By acknowledging these risks, teams can make informed decisions, and allocate resources accordingly.